\pdfoutput=1
\documentclass[journal]{IEEEtran}
\IEEEoverridecommandlockouts
\usepackage{color}
\usepackage{cite}
\usepackage{subfigure}
\usepackage[ruled,vlined,algo2e]{algorithm2e}
\usepackage{algorithm}
\usepackage{algorithmic}
\usepackage{graphicx}
\usepackage{amssymb}
\usepackage{amsmath}
\usepackage[numbers,sort&compress]{natbib}

\usepackage{amsmath}
\begin{document}

\title{Beam Prediction Based on Large Language Models}

\author{Yucheng Sheng,~\IEEEmembership{Student Member,~IEEE,}
        Kai Huang,~\IEEEmembership{Student Member,~IEEE,}
        Le Liang,~\IEEEmembership{Member,~IEEE,}
        \\Peng Liu,~\IEEEmembership{Member,~IEEE,} 
        Shi Jin,~\IEEEmembership{Fellow,~IEEE,}
        Geoffrey Ye Li,~\IEEEmembership{Fellow,~IEEE}
\thanks{Yucheng Sheng, Kai Huang, Le Liang, and Shi Jin are with the National Mobile Communications Research Laboratory, Southeast University, Nanjing 210096, China (e-mail: \{shengyucheng, hkk, lliang, jinshi\}@seu.edu.cn). Le Liang is also with Purple Mountain Laboratories, Nanjing 211111, China.}

\thanks{Peng Liu is with Huawei Technologies Co., Ltd.}
\thanks{Geoffrey Ye Li is with the ITP Lab, the Department of Electrical and Electronic Engineering, Imperial College London, SW7 2BX London, U.K. (e-mail: geoffrey.li@imperial.ac.uk).}
}

\maketitle

\begin{abstract}
In this letter, we use large language models (LLMs) to develop a high-performing and robust beam prediction method. We formulate the millimeter wave (mmWave) beam prediction problem as a time series forecasting task, where the historical observations are aggregated through cross-variable attention and then transformed into text-based representations using a trainable tokenizer. By leveraging the prompt-as-prefix (PaP) technique for contextual enrichment, our method harnesses the power of LLMs to predict future optimal beams. Simulation results demonstrate that our LLM-based approach outperforms traditional learning-based models in prediction accuracy as well as robustness, highlighting the significant potential of LLMs in enhancing wireless communication systems.
\end{abstract}



\begin{IEEEkeywords}
Beam prediction, large language model, time series forecasting, cross attention. 
\end{IEEEkeywords}

\IEEEpeerreviewmaketitle

\section{Introduction}
\IEEEPARstart{M}{illimeter} wave (mmWave) communication is widely acknowledged as a key technology for improving wireless system capacity due to its excessive bandwidth. To address the significant path loss inherent to mmWave signals, large antenna arrays are often deployed at the transmitter and receiver, facilitating directional transmissions. This directional transmission requires beam training to maximize received power. However, the optimal beam direction changes rapidly  in high mobility environments, necessitating frequent beam training and accurate beam prediction.

Deep learning (DL) has recently gained significant research interest in wireless communications due to its strong potential in revolutionalizing communication design principles and dramatically improving communication performance. To improve beam tracking performance, DL is commonly employed to extract UT movement features from received signals \cite{ma2023continuous,cas2021,Li2023attention}, and predict future beam variations. Long short-term memory (LSTM) models, in particular, are typically used to periodically predict the next optimal beam based on signals from previous beam tracking stages \cite{ma2023continuous,cas2021}. However, due to the relatively small number of parameters in LSTM-based models, these models are often sensitive to environment variations, and result in poor robustness and generalization, which is a critical issue for the practical applications of learning-based models. This stands in stark contrast to foundation language models, such as GPT-2 \cite{radford2019language}, GPT-4, and LLaMa \cite{touvron2023llama}, which exhibit remarkable robustness.

Pre-trained foundation models, such as large language models (LLMs), have revolutionized computer vision  and natural language processing fields. Despite some studies suggesting the potential transformative impact of LLMs on wireless communications  \cite{bariah2023large,jiang2023semantic,xie2024towards}, specific applications of these models in beam prediction have yet to be realized. To bridge this gap, we propose a beam prediction model leveraging LLMs, framing the problem as a time series forecasting task. The effectiveness of our approach depends on aligning the modalities of wireless data and natural language. This alignment is challenging because LLMs process discrete tokens, whereas wireless data is analog. Additionally, the ability to interpret wireless data patterns is not inherently included in the pre-training of LLMs. Therefore, it remains an open issue to apply LLMs to process wireless data. 

In this letter, we introduce a framework designed to leverage LLMs for beam prediction without altering the underlying model architecture. Our primary strategy involves aggregating information from diverse variables, which is then transformed into text-based prototype representations to align with the capabilities of language models. To enhance the model's understanding and reasoning of wireless data, we utilize a novel technique, called Prompt-as-Prefix (PaP) \cite{jin2024time}. Simulation results demonstrate that our LLM-based method exhibits superior robustness and generality compared to LSTM-based prediction schemes. Our initial results indicate that LLMs hold promise for diverse applications in wireless communication, shedding lights on the research of LLM-assisted wireless communication.

\section{System Model}


Consider the downlink mmWave transmission for a single user, where the base station (BS) and user terminals (UTs) are equipped with \( M \) antennas and a single antenna, respectively. Our proposed scheme can be directly extended to the multiuser scenario with multiple UT antennas.

We adopt the well-known Saleh-Valenzuela (SV) channel model \cite{saleh1987statistical,liang2014low}, where the channel vector in the \( n \)-th time slot can be expressed as
\begin{equation}
\mathbf{h}_n = \sum_{l=1}^{L_n} \sqrt{\frac{1}{\rho_{n,l}}} \alpha_{n,l} \mathbf{a}^* (\varphi_{n,l}),
\end{equation}
where \( L_n \) denotes the number of paths in the \( n \)-th time slot, while the \( l \)-th path is with path loss \( \rho_{n,l} \), complex gain \( \alpha_{n,l} \), and angle of departure (AoD) \( \varphi_{n,l} \). It should be noted that in this model, the line-of-sight (LOS) path is dominant. Additionally, \( \mathbf{a} \in \mathbb{C}^{M \times 1} \) represents the antenna response vector. Assuming the BS is equipped with a uniform linear array (ULA), the antenna response vector can be expressed as
\begin{equation}
\mathbf{a} (\varphi) = \begin{bmatrix} 1,& e^{j 2 \pi d \sin \varphi / \lambda},& \cdots,&  e^{j \pi (M-1) d \sin \varphi / \lambda} \end{bmatrix}^T,
\end{equation}
where \( \varphi \) is the AoD,  \( d \) denotes the antenna spacing, and \( \lambda \) denotes the wavelength. Without loss of generality, we set \( d = \lambda/2 \).

We assume that a single radio frequency chain and phase shifter based analog beamformer are employed at the BS. The discrete Fourier transform (DFT) codebook \( \mathcal{F} \), comprising \( Q \) candidate beams, is considered. The \( q \)-th candidate transmit beam \( \mathbf{f}^{(q)} \in \mathbb{C}^{M \times 1} \), where \( q \in \{0, 1, 2, \ldots, Q-1\} \), can be expressed as
\begin{equation}
\mathbf{f}^{(q)} = \sqrt{\frac{1}{M}} 
\begin{bmatrix} 
1, & e^{j 2 \pi q / Q}, & \cdots, & e^{j 2 \pi (M-1) q / Q} 
\end{bmatrix}^T.
\end{equation}

Our goal is to predict the optimal beam directions for the next \( H \) time steps based on historical information from the past \( U \) time steps, including the optimal beam and AoD. Mathematically, we aim to find the series of beam indices $\{\hat{q}_1^{\star}, \cdots, \hat{q}_n^{\star}, \cdots\}$ that  achieve the maximum sum of normalized beamforming gain for future $H$ time steps, which can be expressed as
\begin{equation}
\max_{\{\hat{q}_1^{\star},\cdots,\hat{q}_n^{\star},\cdots\}}  \sum_{n=1}^H \frac{|\mathbf{h}_n^{\mathrm{T}} \mathbf{f}^{(\hat{q}_n^{\star})}|^2}{|\mathbf{h}_n^{\mathrm{T}} \mathbf{f}^{(q_n^{\star})}|^2},
\end{equation}
where the optimal beam index in the $n$-th time step is given by
\begin{equation}
q_n^* = \arg\max_{q_n \in \{0, 1, 2, \ldots, Q-1\}} \left| \mathbf{h}_n^\mathrm{T} \mathbf{f}^{(q_n)} \right|^2.
\end{equation}

\section{Prediction Model based on LLMs}
In this letter, we propose reprogramming an embedding-visible LLM, such as Llama and GPT-2, to perform optimal beam forecasting without fine-tuning the backbone model. Specifically, for the \( i \)-th sequence of historical observations \( \mathbf{X}^{(i)} \in \mathbb{R}^{C \times U} \), where $C$ denotes the number of variables and $U$ denotes the length of past observations, we aim to adapt a LLM to interpret the input time series and accurately predict the optimal beam indices for \( H \) future time steps, represented by \( \mathbf{\hat{Y}}^{(i)} \in \mathbb{R}^{1 \times H} \). 

\begin{figure*}[htbp]
    \centering
    \includegraphics[width=1\textwidth]{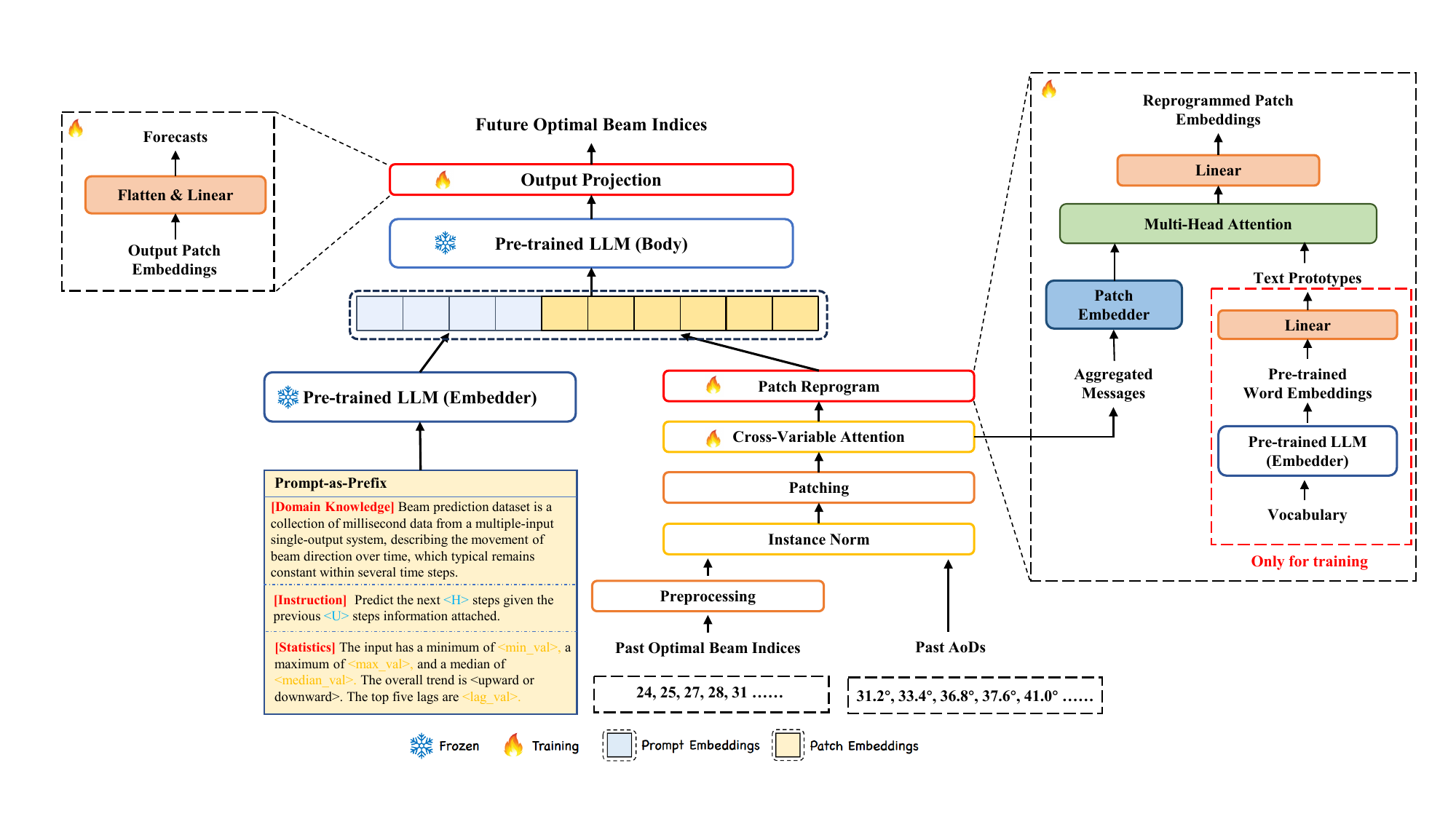}
    \caption{Illustration of LLM-based mmWave prediction. Given the past optimal beam indices and AoDs, we first patch and fuse them using cross-attention. Then after patch reprogramming, they form several discrete tokens together with PaP. The output patches from the LLMs are projected to generate the future optimal beam indices.}
    \label{fig:structure}
\end{figure*}
\textbf{Input Preprocessing.} We use past optimal beam indices $\mathbf{q}^{\star}_s \in \mathbb{R}^{1 \times U}$ and AoD \( \boldsymbol{\phi}_s \in \mathbb{R}^{1 \times U} \) as historical observations. Different numbers of antennas lead to different ranges of beam indices, which reduces generalization. Therefore, we first normalize the sequence of past optimal beam indices \( \mathbf{q}^{\star}_s \), represented by
\begin{equation}
\mathbf{q}^{\star}_a = \frac{\mathbf{q}^{\star}_s}{Q}.
\end{equation}
Finally, we can obtain the input $\mathbf{X}^{(i)}$ by combining $\mathbf{q}^{\star}_a$ and $\boldsymbol{\phi}_s$.

\textbf{Patch Embedding.} Each input sample \( \mathbf{X}^{(i)} \in \mathbb{R}^{C \times U} \) is first normalized to achieve zero mean and unit standard deviation using reversible instance normalization (RevIN). After normalization, \( \mathbf{X}^{(i)} \) is divided into several patches, represented as \( \mathbf{X}_P^{(i)} \), with each patch having a length of \( L_p \). These patches can be either overlapped or non-overlapped, depending on the sliding configuration. The total number of patches is given by
\begin{equation}
P = \left\lfloor \frac{U - L_p}{S} \right\rfloor + 2,
\end{equation}
where \( S \) represents the stride used for horizontal sliding. This segmentation helps retain important local semantic features within each patch and acts as a form of tokenization, effectively creating a compact input sequence that reduces computational complexity for subsequent processing. These patches \( \mathbf{X}_P^{(i)} \in \mathbb{R}^{P \times C \times L_p} \) are then embedded as \( \mathbf{\tilde{X}}_P^{(i)} \in \mathbb{R}^{P \times C \times d_m} \) using a  linear layer as the patch embedding, where \(d_m\) represents the feature dimension. This embedding facilitates effective feature extraction for beam prediction.

\textbf{Cross-Variable Attention.} In this letter, a learnable matrix is applied to aggregate messages from all variables. This is achieved by using $\mathbf{R}^{(i)}$ as the query  and $\mathbf{\tilde{X}}_P^{(i)}$ as both the key and value, denoted by
\begin{equation}
\mathbf{B}^{(i)} = \text{ATTENTION} (\mathbf{R}^{(i)},\mathbf{\tilde{X}}_P^{(i)},\mathbf{\tilde{X}}_P^{(i)}),
\end{equation}
where \( \mathbf{R}^{(i)} \in \mathbb{R}^{P \times 1 \times d_m} \) is a learnable matrix. Cross-variable attention is applied to analyze the relationship between the $\mathbf{q}^{\star}_a$ and \( \boldsymbol{\phi}_s \) and to aggregate messages into \( \mathbf{B}^{(i)} \in \mathbb{R}^{P \times 1 \times d_m} \), which is then passed to patch reprogramming.

\textbf{Patch Reprogramming.} Since LLMs were not pre-trained with time series information, the aggregated feature \( \mathbf{B}^{(i)} \) cannot be directly fed into the LLMs. To address this challenge, we reprogram the aggregated feature \( \mathbf{B}^{(i)} \) into a natural language representation, aligning it with the pre-training input format of LLMs, thereby activating the ability of LLMs to understand and reason about time series data.

Specifically, we reprogram \( \mathbf{B}^{(i)} \) by utilizing pre-trained vocabulary \( \mathbf{E} \in \mathbb{R}^{V \times D} \) from LLMs, where \( V \) represents the vocabulary size and \( D \) represents the hidden dimension of the LLMs. Since directly using \( \mathbf{E} \) would be impractical due to its large vocabulary size, we take a more efficient approach by selecting a reduced set of text prototypes \( \mathbf{E}' \in \mathbb{R}^{V' \times D} \)  from $\mathbf{E}$, where \( V' \ll V \). These text prototypes $\mathbf{E}'$ are designed to capture key language signals, such as ``up" for representing an upward trend or ``down" for describing a downward trend. This approach is efficient and enables an accurate  representation of \( \mathbf{B}^{(i)} \).

To implement reprogramming process, we employ a multi-head cross-attention layer. For each head $k = \{1, \cdots, K\}$, we define query matrices \( \mathbf{Q}_k^{(i)} = \mathbf{B}^{(i)} \mathbf{W}_k^Q \), key matrices \( \mathbf{K}_k^{(i)} = \mathbf{E}' \mathbf{W}_k^K \), and value matrices \( \mathbf{V}_k^{(i)} = \mathbf{E}' \mathbf{W}_k^V \), where \( \mathbf{W}_k^Q \in \mathbb{R}^{d_m \times d} \) and \( \mathbf{W}_k^K, \mathbf{W}_k^V \in \mathbb{R}^{D \times d} \). Specifically, $d = \left\lfloor \frac{d_m}{K} \right\rfloor$ represents the hidden dimension of each head. The reprogramming operation for time series patches in each attention head is then defined as 
\begin{equation}
\mathbf{Z}_k^{(i)} = \text{ATTENTION}(\mathbf{Q}_k^{(i)}, \mathbf{K}_k^{(i)}, \mathbf{V}_k^{(i)}).
\end{equation}

By aggregating each \( \mathbf{Z}_k^{(i)} \in \mathbb{R}^{P \times d} \) in every head, we obtain \( \mathbf{Z}^{(i)} \in \mathbb{R}^{P \times d_m} \), which is then linearly projected to align the hidden dimensions with the backbone model, yielding \( \mathbf{O}^{(i)} \in \mathbb{R}^{P \times D} \).

\textbf{Prompt-as-Prefix.} Recent research suggests that incorporating natural language prompts with other data modalities, such as images, can enhance the reasoning capabilities of LLMs. Similarly, in this paper, we employ prompts as prefixes for the time series information \(\mathbf{O}^{(i)}\), a technique we refer to as Prompt-as-Prefix (PaP), to improve the LLM's adaptability to the beam prediction task. Specifically, we concatenate the prompt embeddings, generated through the pre-trained LLM embedder, with the time series information \(\mathbf{O}^{(i)}\) before feeding them into the pre-trained LLM backbone.

As illustrated in Fig. \ref{fig:structure}, we use three different types of prompts as prefixes, including domain knowledge, instruction, and statistics. As for statistics, we adopt the sum of difference between successive time steps to express the overall trend of the time series in natural language. If the sum is positive, it signifies an upward trend; otherwise, a downward trend. Additionally, we identify the top five lags of the time series by computing the autocorrelation function using fast Fourier transform (FFT) and selecting the five lags with the largest correlation.

\textbf{Output Projection.} After packing and feeding the prompt and patch embeddings \( \mathbf{O}^{(i)} \) through the frozen LLM, as shown in Fig. \ref{fig:structure}, we discard the prefix portion and obtain the output representations. These representations are then flattened and linearly projected to derive the forecasts \( \mathbf{\hat{Y}}^{(i)} \). The forecasts \( \mathbf{\hat{Y}}^{(i)} \) are subsequently converted into future optimal beam indices using a postprocessing step, which reverts the preprocessing procedure. The loss function aims to minimize the mean-squared errors between the ground truth \( \mathbf{Y} \) and the predictions \( \mathbf{\hat{Y}} \) of the optimal beam indices over $H$ future time steps, which can be expressed as
\begin{equation}
L=\frac{1}{H} \sum_{n=1}^{H} \| \mathbf{\hat{Y}}_n - \mathbf{Y} _n \|_2^2.
\end{equation}

\begin{figure}[htbp]
    \centering
    \includegraphics[width=8cm]{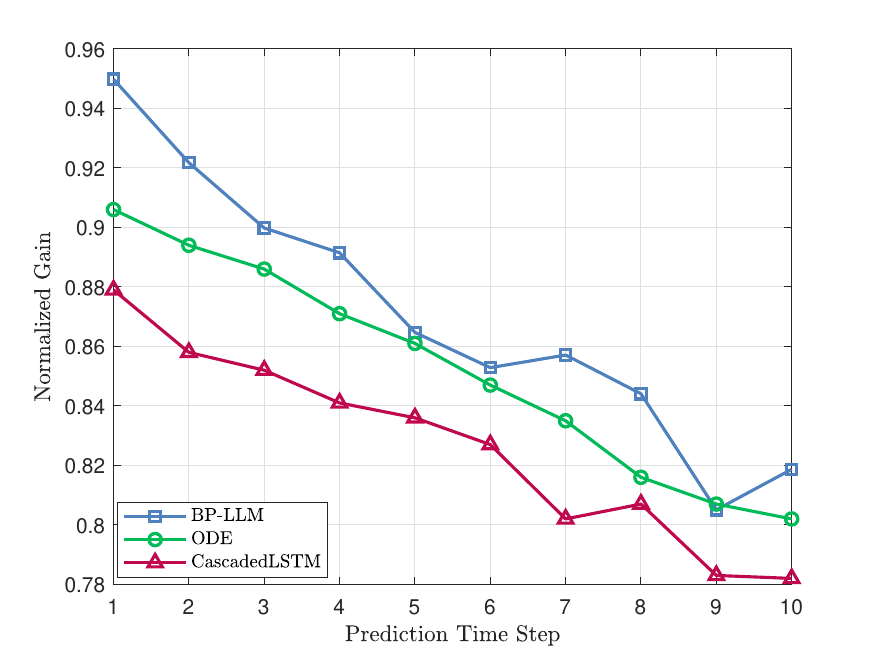}
    \caption{Prediction performance of the proposed method, BP-LLM, compared with other learning-based methods, averaged across test velocities of 5, 10, 15, 20 m/s.}
    \label{fig:v_mismatch}
\end{figure}

\section{Simulation}
\subsection{Simulation system setup}

\begin{figure}[htbp]
    \centering
    \includegraphics[width=8cm]{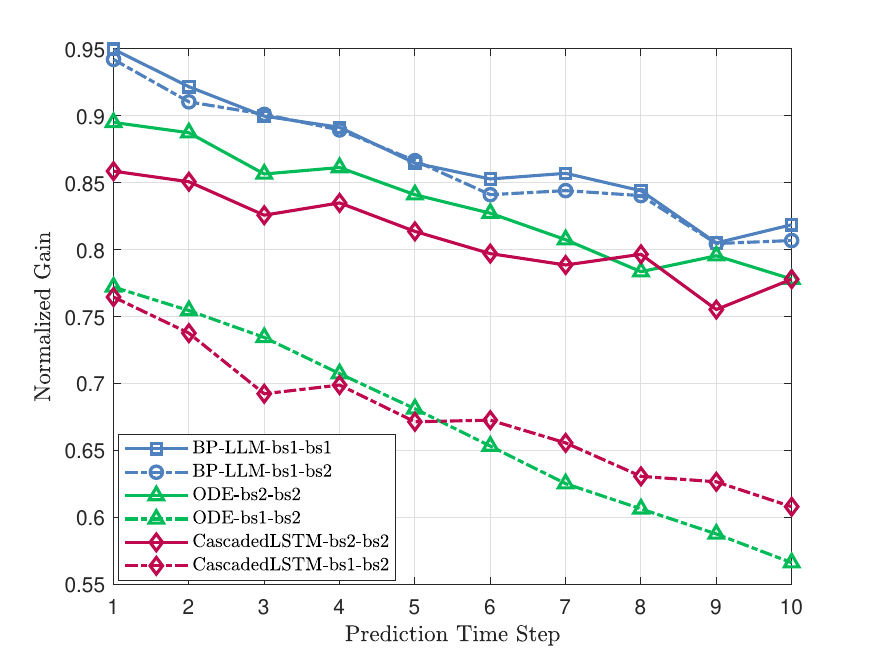}
    \caption{Performance of the proposed method compared with other learning-based methods under the mismatched BS settings.}
    \label{fig:bs_mismatch}
\end{figure}

We select GPT-2 \cite{radford2019language} as our backbone model, which achieves a trade-off between inference speed and prediction accuracy. Note that our method is also theoretically applicable to other LLMs, such as the Qwen and Llama series \cite{touvron2023llama}. As for baselines, we compare our proposed method with two LSTM-based methods, labeled as ODE \cite{ma2023continuous} and CascadedLSTM \cite{cas2021}. Both ODE and CascadedLSTM take the received signal of beam training as their inputs. We adopt the DeepMIMO dataset \cite{deepmimo2019}, which uses precise ray-tracing methods, to collect beam indices and AoDs for training.

A training dataset comprising 73,728 samples and a validation dataset comprising 9,216 samples are constructed, respectively. In these datasets, we focus on an outdoor environment, to reflect real-world implementation. BSs are configured with 32, 64, or 128 antennas, with the number of beams matching the respective antenna numbers to facilitate testing of different communication patterns. Lastly, the period of each beam prediction step is set as 16 ms while $U$ and $H$ are set as 40 and 10 prediction steps, respectively.  The optimal beams computed from the neighborhood of the prediction results will be collected at regular intervals and used as inputs for the next prediction. Note that our model is trained under the assumption that the BS index is set to one with 32, 64, and 128 antennas, the UT velocity ranges between 5 $\sim$ 20 m/s, and the center frequency is set to 28 GHz.

\begin{figure}[htbp]
    \centering
    \includegraphics[width=7.5cm]{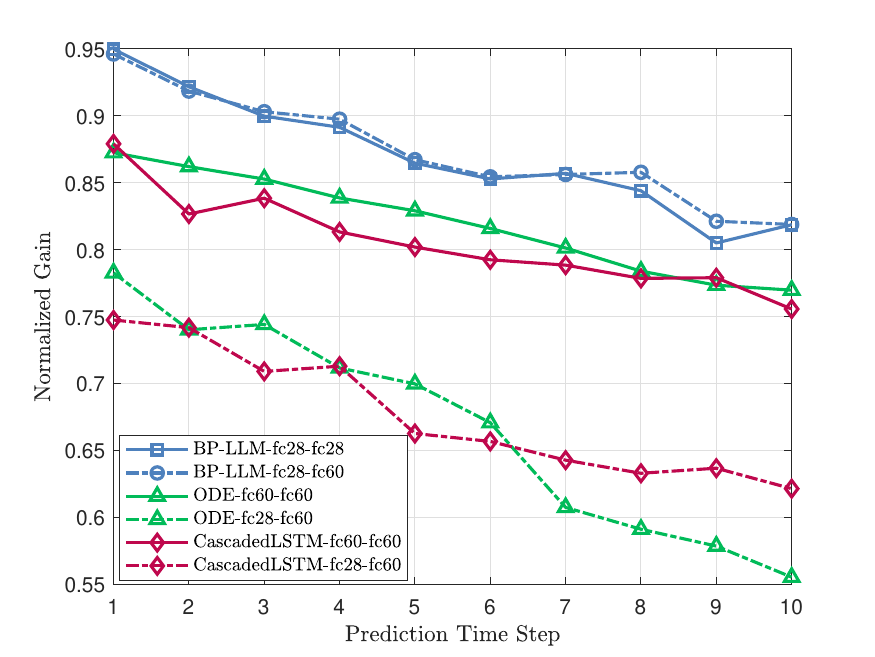}
    \caption{Performance of the proposed method compared with other learning-based methods under the mismatched center frequency.}
    \label{fig:f_mismatch}
\end{figure}

\subsection{Simulation Results}

Fig. \ref{fig:v_mismatch} illustrates the prediction performance of the proposed LLM-based prediction method, labeled as ``BP-LLM", compared with other learning-based methods, which is averaged across testing  velocities of 5, 10, 15, 20 m/s. In Fig. \ref{fig:v_mismatch}, ODE and CascadedLSTM methods are trained at four different speeds and then tested at four different speeds, with the results averaged. The proposed method surpasses both baselines because baseline models, with limited numbers of parameters, lack the robustness needed to effectively handle diverse scenarios. Consequently, training with data from multiple speeds often fails to improve performance and may even lead to a performance degradation. In contrast, LLMs, due to their greater number of parameters and extensive pre-training on large datasets, tend to perform better than traditional smaller models when trained across multiple speeds. Note that beam prediction is not a particularly challenging task, so baseline models already achieve strong performance, making the gains from LLMs appear less significant. However, the additional parameters of LLMs provide enhanced robustness, as demonstrated in the robustness experiments below.

We then evaluate robustness across \textit{different BSs}, as shown in Fig. \ref{fig:bs_mismatch}.  `BP-LLM-bs1-bs2' indicates that the model was trained on the data of BS 1 and tested on the data of BS 2 and the other legends follow the same convention. Fig. \ref{fig:bs_mismatch} indicates that the performance of LSTM-based models significantly deteriorates when tested under the mismatched BS settings. In contrast, our LLM-based solution exhibits strong robustness, with minimal performance drop across different BSs. This demonstrates that our solution leverages the powerful zero-shot learning capabilities of LLMs, enabling them to learn more general domain knowledge even when trained on a single BS. 

We also evaluate robustness across \textit{different center frequencies} in Fig. \ref{fig:f_mismatch}. `BP-LLM-fc28-fc60' indicates that the model was trained on the center frequency of 28 GHz and tested on the center frequency of 60 GHz  and the other legends follow the same convention. Our LLM-based solution demonstrates remarkable robustness compared to the LSTM-based models. It is worthy noting that our robustness stems not only from the large model but also from our input setup. Previous LSTM methods primarily use received signals as input, making them heavily reliant on channel gains that vary with the center frequency. Our solution, which leverages past optimal beam indices and AoDs, is more robust to channel variations.

\begin{figure}[htbp]
    \centering
    \includegraphics[width=8cm]{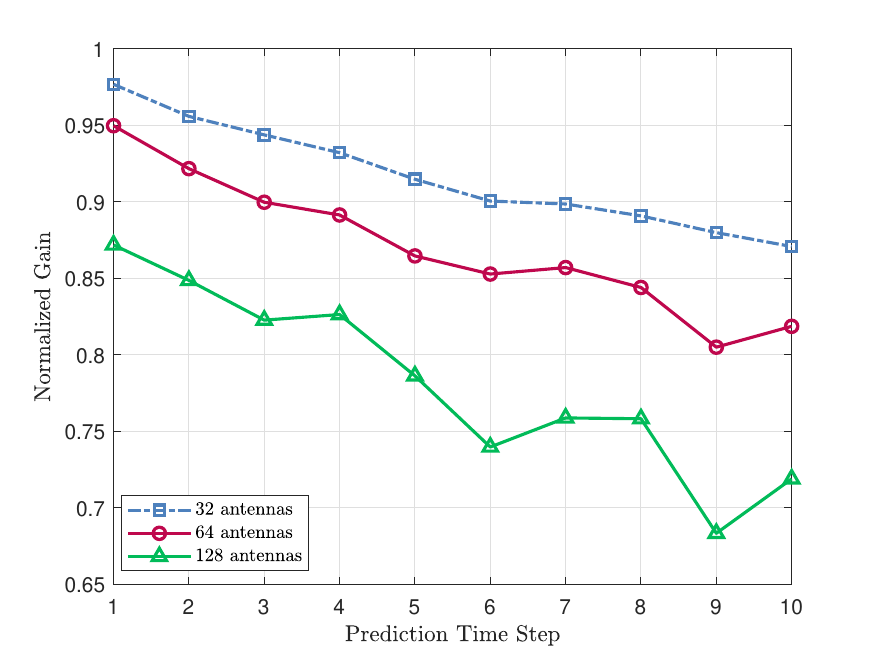}
    \caption{Performance of the proposed system with different antennas.}
    \label{fig:antenna}
\end{figure}
We also conducted tests with \textit{different antenna configurations} in Fig. \ref{fig:antenna}. For previous LSTM-based solutions, their lack of scalability is a significant drawback due to their reliance on received signals as input. This means that the change of antenna numbers necessitate redesign and retraining of previous models. Our solution partially resolves this issue. As shown in Fig. \ref{fig:antenna}, our method can be applied to different antennas without retraining, demonstrating  a notable level of generalization. We also find that as the number of antennas increases, the overall normalized gain decreases. This phenomenon can be attributed to the narrowing beamwidth with more antennas, leading to a significant performance difference between the best beam and other beams.

\begin{figure}[htbp]
    \centering
    \includegraphics[width=8cm]{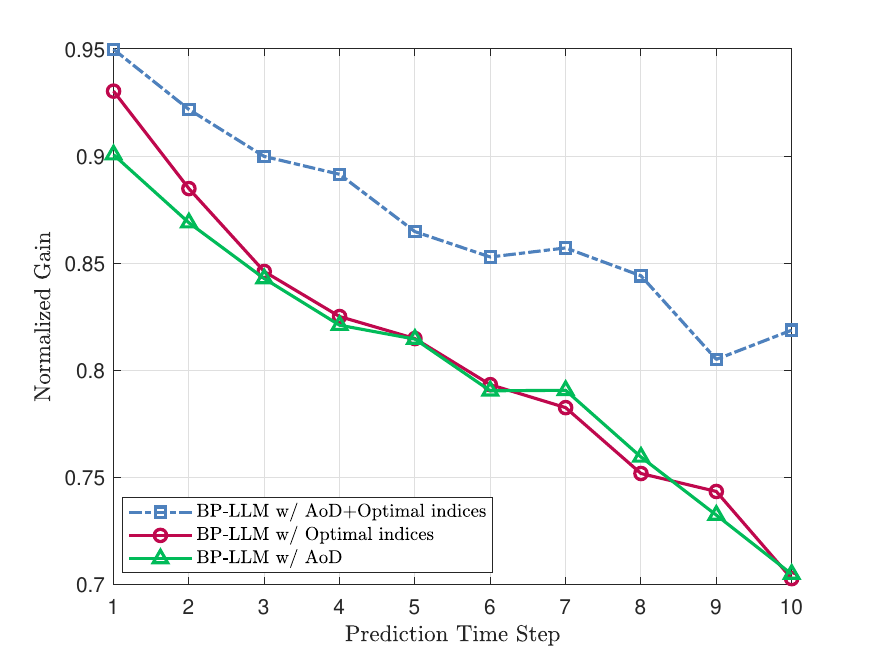}
    \caption{Ablation study on the impacts of different input variables for LLM-based mmWave prediction method. }
    \label{fig:variable}
\end{figure}

\begin{figure}[htbp]
    \centering
    \includegraphics[width=8cm]{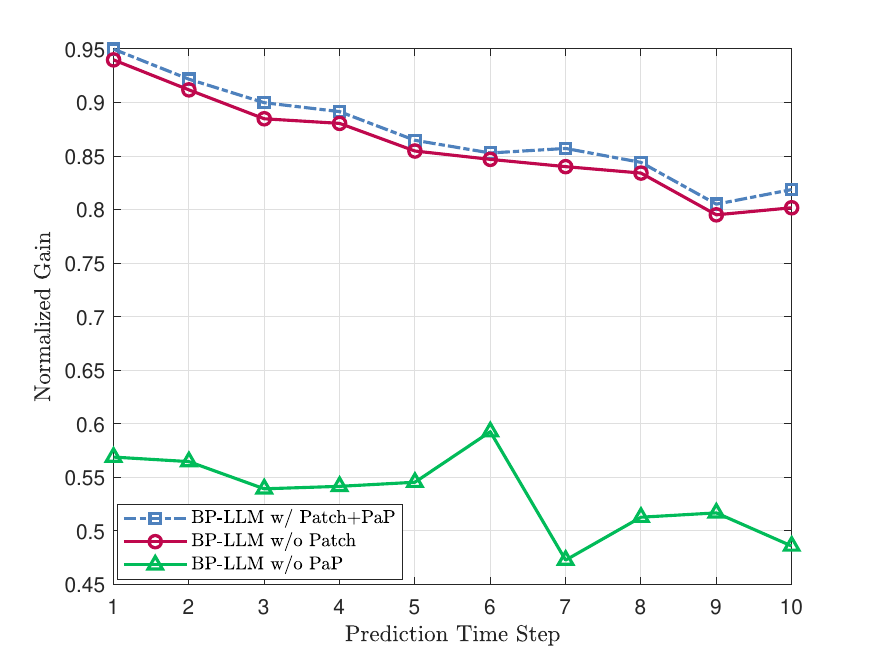}
    \caption{Ablation study on the the impacts of different components of the LLM-based mmWave prediction method. }
    \label{fig:ablation}
\end{figure}

We also conduct an ablation experiment on the different inputs in Fig. \ref{fig:variable}. We compare the schemes of using only one variable (e.g. optimal beam indices or AoD) as input versus using the optimal beam indices plus AoD. We find that combining  the optimal beam indices with the AoD  information significantly improves performance. Lastly, we conduct an ablation study on the different components of the proposed model in Fig.~\ref{fig:ablation}. We find that our method without patch embedding (`BP-LLM w/o Patch') results in only a minor performance drop, whereas removing PaP (`BP-LLM w/o PaP') leads to significant degradation. This demonstrates that PaP plays a crucial role in enhancing the ability of LLMs to effectively understand beam and AoD information.

\section{Conclusion}
In this paper, we have presented a novel framework that adapts LLM for beam prediction. By aggregating the optimal beam indices and AoD, and subsequently converting them into text-based prototype representations, we align the data format with LLM capabilities. Our innovative PaP technique further enhances the model’s understanding and reasoning of wireless data.  Our findings suggest that LLMs have significant potential for diverse applications in wireless communications.

\small
\bibliographystyle{IEEEtran}
\bibliography{journal_dc.bib}

\end{document}